# A Region-Based Deep Learning Approach to Automated Retail Checkout

Maged Shoman\*& Armstrong Aboah†
Department of Civil Engineering
University of Missouri-Columbia
mas5nh, aa5mv@missouri.edu

Alex Morehead & Ye Duan
Department of EECS
University of Missouri-Columbia
acmwhb, duanye@missouri.edu

Abdulateef Daud & Yaw Adu-Gyamfi
Department of Civil Engineering
University of Missouri-Columbia
aadcvg, adugyamfiy@missouri.edu

## Abstract

*Automating the product checkout process at conventional retail stores is a task poised to have large impacts on society generally speaking. Towards this end, reliable deep learning models that enable automated product counting for fast customer checkout can make this goal a reality. In this work, we propose a novel, region-based deep learning approach to automate product counting using a customized YOLOv5 object detection pipeline and the DeepSORT algorithm. Our results on challenging, real-world test videos demonstrate that our method can generalize its predictions to a sufficient level of accuracy and with a fast enough runtime to warrant deployment to real-world commercial settings. Our proposed method won 4th place in the 2022 AI City Challenge, Track 4, with an F1 score of 0.4400 on experimental validation data.*

## 1. Introduction

Retail checkout is a process that influences much of our everyday lives, whether or not we consciously give this process much thought. Traditionally, the checkout procedure in commercial settings such as grocery stores has been performed for customers in-person by store associates. Today, there is a rising level of interest in seeing these kinds of tasks automated and relinquishing these procedures from associates who would traditionally carry them out, to enable such associates to invest their time and skillsets into more specialized roles. Deep learning, an ever-expanding subfield of machine learning, aims to learn arbitrary functions using parameterized systems often referred to as neu-ral networks, for increasingly complex tasks such as computer vision and natural language processing [30-35]. In this work, we thoroughly explore the use of deep learning in a commercial retail setting by proposing a scalable computer vision method that enables automated product counting and, subsequently, automatic product checkout *en masse*.

In this study, we developed a framework specifically for automatic retail checkout. The proposed methodology relies on first building a robust object detection model using YOLOv5 [29]. Next, our pipeline identifies a region of interest (ROI) in every video by initially estimating the background of the video (i.e., computing the median of frames randomly sampled from a uniform distribution over the entire duration of the video), followed by ROI identification using adaptive thresholding. A selected ROI is then passed through a custom-trained YOLOv5 model for detection. The detections made within the ROI are further tracked using the DeepSORT algorithm. Finally, the time an object is first detected within the ROI is computed by finding the ratio of the frame number to the video frequency rate, thereby giving us precise time measurements of an object's first sighting within the ROI.

Our experimental results demonstrate that our proposed method can generalize its predictions to account for many of the visual complexities and situational ambiguities that often occur in a retail checkout space. Furthermore, they demonstrate that the runtime of our proposed method is acceptable for a retail context, maintaining high throughput when analyzing input video sequences.

The rest of the paper is organized as follows. Section 2 provides a review of the relevant literature. The data used for this study is described in Section 3. Section 4 presents the methodology introduced by our study. Section 5 presents a discussion of the results from our model's development. Finally, Section 6 presents a summary of our research, the conclusions drawn from our results, and our recommendations for future research.

---

\*Equal contribution.
†Equal contribution.

## 2. Related Work

An increasing number of researchers have begun investigating the use of machine learning systems in retail environments [5]. For example, [18] describe specific technical outcomes needed for retail spaces to fully adopt automated retail practices such as fully self-service product checkout. [7] details the challenges associated with fully implementing a self-service checkout system in retail spaces. [13] outlines the societal impacts that machine learning can have on how consumers shop for products in retail spaces.

From the perspective of classical computer vision methodologies in retail spaces, [15] present a method that employs motion segmentation, template matching, and line detection to ensure that customers' shopping carts are demonstrably empty during cart compliance checks. To assist visually-impaired shoppers in locating stocked items of interest, [25] train a Naive-Bayes classifier [22] on image features extracted using the SURF algorithm [1].

The supermarket checkout procedure, as many of us are already aware, is time-consuming, often requires human supervision, and typically involves lengthy waiting lines. To address this, [3] proposed an automatic retail checkout (ARC) system using a computer vision-based system that scans objects placed beneath a webcam for object identification. This method aims to make the process of checkout at retail stores faster, more convenient, and less dependent on supervisory intervention. The promise of ARC is that it would be fully autonomous, helping retail stores reduce their reliance on human operators for repetitive tasks such as product checkout.

To date, relatively few works have directly addressed the task of automated product checkout from the perspective of contemporary computer vision systems driven by deep learning. [8] introduces a hypothetical object detection pipeline for retail checkout, based on the popular YOLOv4 algorithm [2]. Subsequently, [21] in a detailed survey investigated and tested several computer vision algorithms such as Mask R-CNN [6] and YOLOv3 [20] in a simulated retail self-checkout environment. These studies highlighted the concerns for balancing model accuracy in product counting as well as the runtime efficiency of the overall algorithm.

From the technical perspective of computer vision in retail contexts, object detection and object tracking are some of the most important computer vision applications, as they involve detecting and classifying objects that are present in a particular image or video. [11]. Notably, object classifiers have been re-purposed to perform detections in previous work [17]. [19] introduced a new approach by framing object detection as a regression problem targeting the spatial generation of bounding boxes and their associated class probabilities. The YOLO algorithm [19] they proposed is trained using a loss function that directly relates to detection performance, while a complete object classification model is learned concurrently.

Similarly, object trackers have recently seen a rise in adoption in large part due to advances in neural networks designed for object tracking [4]. [10] survey the rise in popularity of using deep convolutional and attentional neural networks for automatic object tracking. Concretely, [23] propose Deep Affinity Networks for multi-object tracking and demonstrate the advantages of using deep neural networks for general object tracking.

Multi-object tracking poses several obstacles, including similarity and high density of detected items, as well as occlusions and perspective shifts while the object moves. [12] attempted to address these challenges by initializing IDs for each object and considering an object as tracked if it has been detected in previous frames and passing this tracking information to the remainder of the tracking framework.

Particularly interesting to note, [28] showcase using the YOLOv5 algorithm with Transformer [24] heads for robust object detection. As we will describe further in subsequent sections, the original YOLOv5 object detection algorithm is a suitable deep learning method for computer vision in realworld settings, achieving competitive performance at a fraction of the runtime requirements of comparable computer vision systems while maintaining easy integration with deep object tracking algorithms such as DeepSORT [26]. As DeepSORT builds upon the popular SORT object detection algorithm using learned convolution appearance descriptors, integrating YOLOv5 and DeepSORT into an endto-end object detection and tracking pipeline holds distinct advantages in terms of information flow, an insight that we exploit thoroughly in this work.

## 3. Data

### 3.1. Data Overview

The data consists of a large set of 116,500 synthetic and real images, with their respective masks belonging to 116 classes. The classes are traditional retail objects that can be found at a retail store, as seen in Figure 1a. We focus on detecting 116 different foreground object instances with a wide range of colors, textures (e.g., homogeneous colors vs. heavy textures), 3D shapes, and material attributes (e.g., reflective vs. non-reflective). A sample of the various images per class is presented in Figures 1a and 1b.

### 3.2. Data Processing

In order to build a robust object detection model, one of the major steps taken in this study was performing certain key data pre-processing steps. Figure 2 presents an overview of the pre-processing steps taken in this study. We performed two major data pre-processing steps to achieve great results with our object detection model. First, we carried out augmentation techniques to increase the diversity of our training data. Second, we extracted bounding boxes from the segmentation mask provided for each object in our training dataset. Figure 3 presents a visual sample of our data pre-processing strategy, one that incorporates the following four data augmentation techniques:

- Mosaic data enhancement combines multiple images or objects into a single large image.
- CutMix clips out a portion of the image and then flips the cropped region with another image.
- Blur reduces the sharpness of an image by smoothing the color transition between pixels.
- Lastly, geometric distortion introduces random scaling, cropping, flipping, and rotating.

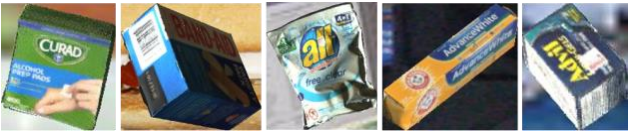

(a) A selection of multi-class objects.

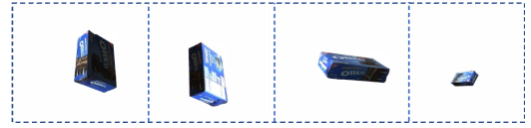

(b) Variations of intra-class images.

Figure 1. Sample images from the training dataset.

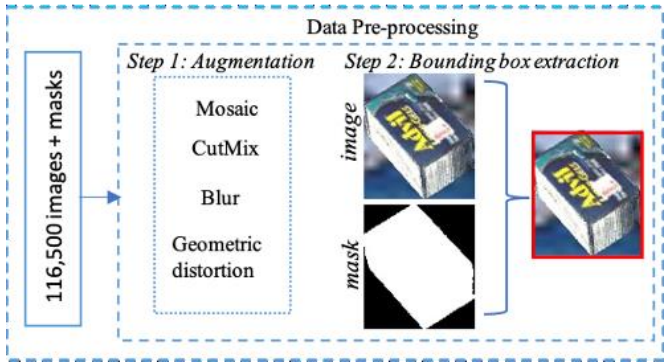

Figure 2. An overview of our data pre-processing scheme.

## 4. Methodology

The challenge is formulated as an object recognition and tracking problem. Our approach is centered around three main ideas: identification of an ROI, fast object detection, and deep object tracking. An overview of our multi-class product counting and recognition pipeline is presented in Figure 4. The first step in our pipeline is to identify an ROI through an adaptive thresholding technique. Next, we perform object detection in the ROI using a custom-trained YOLOv5, which is then followed by deep object tracking with DeepSORT.

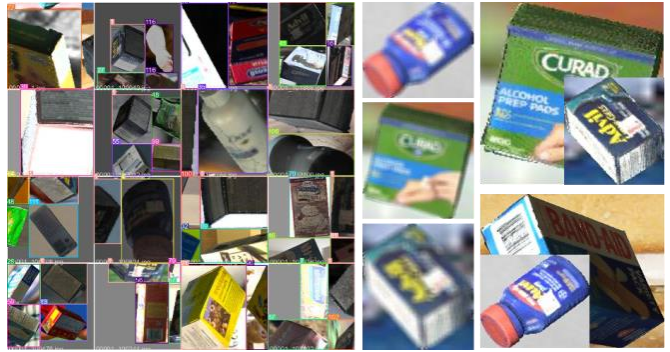

Figure 3. Augmentations performed on our training images: mosaic (*left*) for combining images, blur (*middle*) to reduce sharpness of an image, CutMix (*right*) to flip cropped region of an image with another image.

### 4.1. Identification of an ROI

A problem-specific observation to be made for automated retail checkout is that the area of focus for an object detector designed for this task is, by definition, in a narrow range. As shown in Figure 4, we leverage this insight within our detection and tracking pipeline. The background for each test video is estimated by randomly sampling frames within the video's length and computing the median of 10% of all frames. Random sampling and the use of this median on a subset of the background images eliminates the effect of short-term video resolution changes such as zoom, pixelation, and moving objects. We then used an adaptive image thresholding technique to generate masks from the background image to identify a consistent ROI throughout all testing videos. We employed Equation 1 for background image thresholding.

$$(\mu - K_1(\sigma))/K_2 \leq T \leq (\mu + K_1(\sigma))/(K_1 + K_2) \quad (1)$$

Here, $K_1$ and $K_2$ are selected based on the pixel histogram distribution of our testing videos. Such a design decision allows our object detector to focus on correctly classifying and detecting objects within the described ROI, which in our case aligns with the checkout tray present in our test dataset's checkout videos for model evaluation.

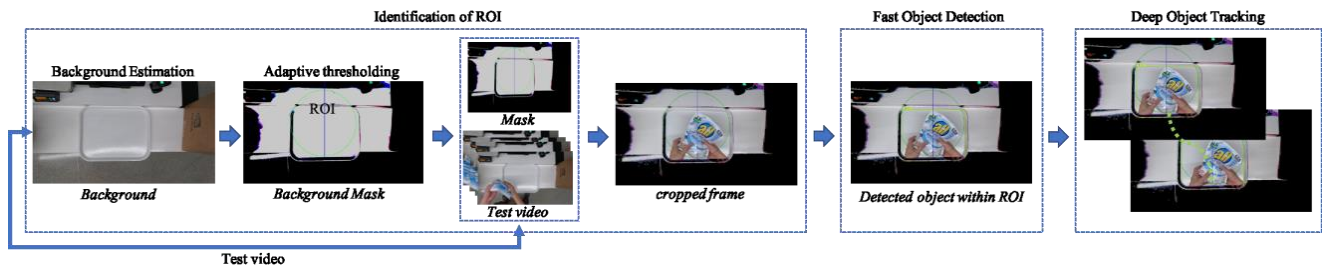

Figure 4. A visual inductive bias we exploit within our pipeline for automated retail checkout.

### 4.2. Fast Object Detection

Our methodology for object detection, in this work, centers around the new YOLOv5 algorithm [9] for object classification and detection. The YOLO object recognition model was the first to combine bounding box estimates and object classification in an end-to-end differentiable network. Notably, Darknet is the setting in which YOLO was developed and is maintained.

Relevant to our task, YOLOv5 is the first YOLO model built using the PyTorch framework [16] and, as such, is much faster and simpler to use than previous YOLO models. For real-time object detection, YOLOv5 uses a robust Convolutional Neural Network (CNN). This algorithm divides the input image into regions and then calculates for each region bounding boxes and probabilities. These bounding boxes are then weighted based on the model's estimated probabilities.

YOLOv5 pushes forward the state-of-the-art by incorporating ideas such as weighted residual connections, crossstage partial connections, cross mini-batches, normalization, and self-adversarial training. To produce predictions, the method only needs one forward pass through the neural network, so it *only looks once* at the image. After applying non-maximal suppression [14], it outputs known objects along with their bounding boxes. Doing so ensures that the object detection algorithm only recognizes each object once. Such a model allows for accurate and swift object detection in real-time, with easy extensibility for task specificity.

### 4.3. Deep Object Tracking

Once we have obtained object detections from our trained YOLOv5 algorithm, we then employ a customized version of the DeepSORT algorithm [26] to track the detected objects within the identified ROI. Object tracking is performed to ensure that objects are counted accurately. The DeepSORT algorithm [27] incorporates appearance information about items in order to improve its frame-to-frame object associations. The object association procedure incorporates an extra appearance parameter based on pretrained CNNs that enables the re-identification of objects during an extended period of occlusion during our detection process.

Finally, using outputs provided by DeepSORT, we divided the frame number of an object's first appearance by the frequency rate (i.e., frames per second) of the respective video to obtain the time at which the object was first detected in the ROI.

## 5. Results and Discussion

Task 4 of the 2022 NVIDIA AI City Challenge presents a total of 116,500 synthetic images with their respective masks for training and 5 videos of real-world checkout scenarios for testing. Each testing video for this task is less than 60 seconds and has a resolution of 1920 × 1080 pixels. The first phase of this task, shown in Figure 4 (Background), presents a frame from the testing video where the camera is mounted above the checkout counter. Models designed for this task should focus on the shopping tray, and items should be detected by their class once they are in front of the tray. A sample of the detections performed on the testing videos using our trained model is presented in Figure 5.

A submission to the online evaluation system for this challenge is one text file in the following format: Video ID, Class ID, Timestamp. Video ID is a video's numeric identifier, starting with 1. It represents the position of the video in the list of all Track 4, Test Set A videos, sorted in alphanumeric order. Class ID is the object numeric identifier, starting with 1. Lastly, Timestamp is the time in the video when the object was first identified, in seconds. The timestamp is an integer and represents a time when the item is within the ROI (i.e., over the white tray). Each object should be identified only once when it passes through the ROI. An example submission of ours is shown in Table 1.

Evaluation is based on the F1 score (i.e., Equation 2) where a true positive (*TP*) identification is counted when an object was correctly identified within the ROI (i.e., the object class was correctly determined) *and* the object was identified within the time that the object was in front of the white tray. A false positive (*FP*) is an identified object that is not a *TP* identification. Finally, a false negative (*FN*) identification is a ground-truth object that was not correctly identified.

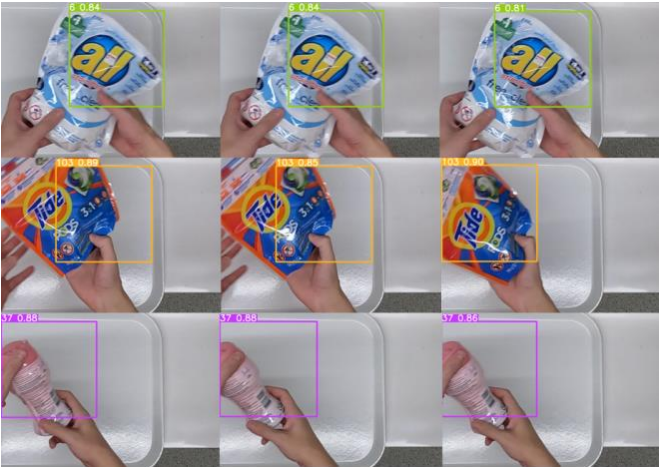

Figure 5. Samples of detected items. Zoomed-in pictures display class IDs and confidence levels.

Table 1. Sample of submitted results.

| Video ID | Class ID | Timestamp |
|---|---|---|
| 1 | 6 | 76 |
| 1 | 76 | 15 |
| 1 | 106 | 19 |

$$F1 = \frac{TP}{TP + \frac{1}{2}(FP + FN)} \qquad (2)$$

We evaluated our model's performance on the 2022 AI City Challenge's Task 4 test dataset. As shown in Table 1, its F1 score on this dataset is 0.4400, demonstrating the effectiveness and robustness of our method in comparison to other participants in the challenge. The leaderboard for the top 10 teams is also shown in Table 1. As these results demonstrate, our model's overall performance is ranked in 4th place for this task.

## 6. Conclusions

In this work, we explored automating product checkout at retail stores using a novel YOLOv5 and DeepSORT object tracking pipeline. Our proposed method is lightweight, fast, and robust when dealing with diverse images, object shapes, and noise. Our proposed system, having been validated on challenging real-world video data, is poised to have immediate positive impacts on society by enabling fully-autonomous product checkout. In the future, we expect to see advancements in the degree of accuracy allowed by such a system in real-world settings, along with larger training and validation datasets for model evaluation.

Table 2. Final Rankings for the 2022 AI City Challenge, Task 4.

| Rank | Team ID | Team Name | Score |
|---|---|---|---|
| 1 | 16 | BUPT-MCPRL2 | 1.00000 |
| 2 | 94 | SKKU Automation Lab | 0.4783 |
| 3 | 104 | The Nabeelians | 0.4545 |
| 4 | 165 | Mizzou | 0.4400 |
| 5 | 66 | RongRongXue | 0.4314 |
| 6 | 76 | Starwar | 0.4231 |
| 7 | 117 | GRAPH@FIT | 0.4167 |
| 8 | 4 | HCMIU-CVIP | 0.4082 |
| 9 | 9 | CyberCore-Track4 | 0.4000 |
| 10 | 55 | UTE-AI | 0.4000 |

## References

[1] Herbert Bay, Andreas Ess, Tinne Tuytelaars, and Luc Van Gool. Speeded-up robust features (surf). *Comput. Vis. Image Underst.*, 110(3):346–359, jun 2008. 2

[2] Alexey Bochkovskiy, Chien-Yao Wang, and Hong-Yuan Mark Liao. Yolov4: Optimal speed and accuracy of object detection. *ArXiv*, abs/2004.10934, 2020. 2

[3] Syed Talha Bukhari, Abdul Wahab Amin, Muhammad Abdullah Naveed, and Muhammad Rzi Abbas. Arc: A vision based automatic retail checkout system. *arXiv preprint arXiv:2104.02832*, 2021. 2

[4] Gioele Ciaparrone, Francisco Luque Sánchez, Siham Tabik, Luigi Troiano, Roberto Tagliaferri, and Francisco Herrera. Deep learning in video multi-object tracking: A survey. *Neurocomputing*, 381:61–88, 2020. 2

[5] Matthias Hauser, Sebastian A Günther, Christoph M Flath, and Fréderic Thiesse. Towards digital transformation in fashion retailing: A design-oriented is research study of automated checkout systems. *Business & Information Systems Engineering*, 61(1):51–66, 2019. 2

[6] Kaiming He, Georgia Gkioxari, Piotr Dollár, and Ross Girshick. Mask r-cnn. In *Proceedings of the IEEE international conference on computer vision*, pages 2961–2969, 2017. 2

[7] Rabab Alayham Abbas Helmi, Alvin Ti Lin Lee, Md Gapar Md Johar, Arshad Jamal, and Liew Fong Sim. Quantum checkout: An improved smart cashier-less store checkout counter system with object recognition. In *2021 IEEE 11th IEEE Symposium on Computer Applications & Industrial Electronics (ISCAIE)*, pages 151–156. IEEE, 2021. 2

[8] Namitha James, Nikhitha Theresa Antony, Sara Philo Shaji, Sherin Baby, and Jyotsna Annakutty. Automated checkout for stores: A computer vision approach. *RE-*

*VISTA GEINTEC-GESTAO INOVACAO E TECNOLOGIAS*, 11(3):1830–1841, 2021. 2

[9] Glenn Jocher, Ayush Chaurasia, Alex Stoken, Jirka Borovec, NanoCode012, Yonghye Kwon, TaoXie, Jiacong Fang, imyhxy, Kalen Michael, Lorna, Abhiram V, Diego Montes, Jebastin Nadar, Laughing, tkianai, yxNONG, Piotr Skalski, Zhiqiang Wang, Adam Hogan, Cristi Fati, Lorenzo Mammana, AlexWang1900, Deep Patel, Ding Yiwei, Felix You, Jan Hajek, Laurentiu Diaconu, and Mai Thanh Minh. ultralytics/yolov5: v6.1 - TensorRT, TensorFlow Edge TPU and OpenVINO Export and Inference, Feb. 2022. 4

[10] Lesole Kalake, Wanggen Wan, and Li Hou. Analysis based on recent deep learning approaches applied in real-time multi-object tracking: A review. *IEEE Access*, 9:32650–32671, 2021. 2

[11] Yang Liu, Peng Sun, Nickolas Wergeles, and Yi Shang. A survey and performance evaluation of deep learning methods for small object detection. *Expert Systems with Applications*, 172:114602, 2021. 2

[12] Dimitrios Meimetis, Ioannis Daramouskas, Isidoros Perikos, and Ioannis Hatzilygeroudis. Real-time multiple object tracking using deep learning methods. *Neural Computing and Applications*, pages 1–30, 2021. 2

[13] Simon Moore, Sandy Bulmer, and Jonathan Elms. The social significance of ai in retail on customer experience and shopping practices. *Journal of Retailing and Consumer Services*, 64:102755, 2022. 2

[14] Alexander Neubeck and Luc Van Gool. Efficient non maximum suppression. In *18th International Conference on Pattern Recognition (ICPR'06)*, volume 3, pages 850–855. IEEE, 2006. 4

[15] Unsang Park, Charles A. Otto, and Sharath Pankanti. Cart auditor: A compliance and training tool for cashiers at checkout. In *2010 Fourth Pacific-Rim Symposium on Image and Video Technology*, pages 151–155, 2010. 2

[16] Adam Paszke, Sam Gross, Francisco Massa, Adam Lerer, James Bradbury, Gregory Chanan, Trevor Killeen, Zeming Lin, Natalia Gimelshein, Luca Antiga, et al. Pytorch: An imperative style, high-performance deep learning library. *Advances in neural information processing systems*, 32, 2019. 4

[17] Francisco Pérez-Hernández, Siham Tabik, Alberto Lamas, Roberto Olmos, Hamido Fujita, and Francisco Herrera. Object detection binary classifiers methodology based on deep learning to identify small objects handled similarly: Application in video surveillance. *Knowledge-Based Systems*, 194:105590, 2020. 2

[18] Rajasshrie Pillai, Brijesh Sivathanu, and Yogesh K Dwivedi. Shopping intention at ai-powered automated retail stores (aipars). *Journal of Retailing and Consumer Services*, 57:102207, 2020. 2

[19] Joseph Redmon, Santosh Divvala, Ross Girshick, and Ali Farhadi. You only look once: Unified, real-time object detection. In *Proceedings of the IEEE conference on computer vision and pattern recognition*, pages 779–788, 2016. 2

[20] Joseph Redmon and Ali Farhadi. Yolov3: An incremental improvement. *arXiv preprint arXiv:1804.02767*, 2018. 2

[21] Anton Rigner. Ai-based machine vision for retail self checkout system. *Master's Theses in Mathematical Sciences*, 2019. 2

[22] Irina Rish et al. An empirical study of the naive bayes classifier. In *IJCAI 2001 workshop on empirical methods in artificial intelligence*, volume 3, pages 41–46, 2001. 2

[23] ShiJie Sun, Naveed Akhtar, HuanSheng Song, Ajmal Mian, and Mubarak Shah. Deep affinity network for multiple object tracking. *IEEE Transactions on Pattern Analysis and Machine Intelligence*, 43(1):104–119, 2021. 2

[24] Ashish Vaswani, Noam Shazeer, Niki Parmar, Jakob Uszkoreit, Llion Jones, Aidan N Gomez, Łukasz Kaiser, and Illia Polosukhin. Attention is all you need. *Advances in neural information processing systems*, 30, 2017. 2

[25] Tess Winlock, Eric Christiansen, and Serge Belongie. Toward real-time grocery detection for the visually impaired. In *2010 IEEE Computer Society Conference on Computer Vision and Pattern Recognition - Workshops*, pages 49–56, 2010. 2

[26] Nicolai Wojke, Alex Bewley, and Dietrich Paulus. Simple online and realtime tracking with a deep association metric. *CoRR*, abs/1703.07402, 2017. 2, 4

[27] Nicolai Wojke, Alex Bewley, and Dietrich Paulus. Simple online and realtime tracking with a deep association metric. *2017 IEEE International Conference on Image Processing (ICIP)*, pages 3645–3649, 2017. 4

[28] Xingkui Zhu, Shuchang Lyu, Xu Wang, and Qi Zhao. Tphyolov5: Improved yolov5 based on transformer prediction head for object detection on drone-captured scenarios. In *Proceedings of the IEEE/CVF International Conference on Computer Vision*, pages 2778–2788, 2021. 2

[29] Aboah, A. (2021). A vision-based system for traffic anomaly detection using deep learning and decision trees. In *Proceedings of the IEEE/CVF Conference on Computer Vision and Pattern Recognition* (pp. 4207-4212).

[30] Shoman, M., Aboah, A., & Adu-Gyamfi, Y. (2020). Deep learning framework for predicting bus delays on multiple routes using heterogenous datasets. *Journal of Big Data Analytics in Transportation*, *2*(3), 275-290.

[31] Aboah, A., & Adu-Gyamfi, Y. (2020). Smartphone-Based Pavement Roughness Estimation Using Deep Learning with Entity Embedding. *Advances in Data Science and Adaptive Analysis*, *12*(03n04), 2050007.

[32] Shoman, M., Amo-Boateng, M., & Adu-Gyamfi, Y. Multi-Purpose, Multi-Step Deep Learning Framework for Network-Level Traffic Flow Prediction. *Available at SSRN 4063434*.

[33] Aboah, A., Boeding, M., & Adu-Gyamfi, Y. (2021). Mobile Sensing for Multipurpose Applications in Transportation. *arXiv preprint arXiv:2106.10733*.

[34] Dadzie, E., & Kwakye, K. (2021). Developing a Machine Learning Algorithm-Based Classification Models for the Detection of High-Energy Gamma Particles. *arXiv preprint arXiv:2111.09496*.

[35] Kwakye, K., Seong, Y., & Yi, S. (2020, August). An Android-based mobile paratransit application for vulnerable road users. In *Proceedings of the 24th Symposium on International Database Engineering & Applications* (pp. 1-5).